\documentclass[conference, 9pt]{IEEEtran}
\IEEEoverridecommandlockouts
\usepackage{cite}
\usepackage{amsmath,amssymb,amsfonts}
\usepackage{algorithmic}
\usepackage{graphicx}
\usepackage{textcomp}
\usepackage{xcolor}
\usepackage{microtype}

\usepackage {inputenc}
\usepackage{adjustbox}
\usepackage{multirow}
\usepackage{todonotes}
\PassOptionsToPackage{hyphens}{url}\usepackage{hyperref}
\usepackage[normalem]{ulem}
\useunder{\uline}{\ul}{}

\usepackage{booktabs,makecell,tabularx}

\newcolumntype{C}[1]{>{\centering\arraybackslash}p{#1}}
\newcolumntype{L}{>{\raggedright\arraybackslash}X}

\usepackage{siunitx}
\usepackage{etoolbox}
\newrobustcmd{\B}{\bfseries}

\newcommand{\rparagraph}[1]{\vspace{1.4mm}\noindent\textbf{#1.}}

\newcommand{\sparagraph}[1]{\vspace{0.0mm}\noindent\textbf{#1.}}

\def\BibTeX{{\rm B\kern-.05em{\sc i\kern-.025em b}\kern-.08em
    T\kern-.1667em\lower.7ex\hbox{E}\kern-.125emX}}
\begin{document}
\title{SpeechTaxi: On Multilingual Semantic Speech Classification \\
}

\author{\IEEEauthorblockN{Lennart Keller}
\IEEEauthorblockA{\textit{Center For Artificial Intelligence and Data Science, } \\
\textit{University of Würzburg, }\\
Würzburg, Germany \\
lennart.keller@uni-wuerzburg.de}
\and
\IEEEauthorblockN{Goran Glavaš}
\IEEEauthorblockA{\textit{Center For Artificial Intelligence and Data Science, } \\
\textit{University of Würzburg, }\\
Würzburg, Germany \\
goran.glavas@uni-wuerzburg.de}
}

\maketitle

\begin{abstract}
Recent advancements in multilingual speech encoding as well as transcription raise the question of the most effective approach to semantic speech classification. Concretely, can (1) end-to-end (E2E) classifiers obtained by fine-tuning state-of-the-art multilingual speech encoders (MSEs) match or surpass the performance of (2) cascading (CA), where speech is first transcribed into text and classification is delegated to a text-based classifier.
To answer this, we first construct SpeechTaxi, an 80-hour multilingual dataset for semantic speech classification of Bible verses, covering 28 diverse languages.   
We then leverage SpeechTaxi to conduct a wide range of experiments comparing E2E and CA in monolingual semantic speech classification as well as in cross-lingual transfer. 
We find that E2E based on MSEs outperforms CA in monolingual setups, i.e., when trained on in-language data.
However, MSEs seem to have poor cross-lingual transfer abilities, with E2E substantially lagging CA both in (1) zero-shot transfer to languages unseen in training and (2) multilingual training, i.e., joint training on multiple languages. Finally, we devise a novel CA approach based on transcription to Romanized text as a language-agnostic intermediate representation and show that it represents a robust solution for languages without native ASR support.
Our SpeechTaxi dataset is publicly available at: \url{https://huggingface.co/datasets/LennartKeller/SpeechTaxi/}.
\end{abstract}

\begin{IEEEkeywords}
Speech Classification, Multilingual Dataset, Spoken Language Understanding, Transliteration, Cross-Lingual Transfer
\end{IEEEkeywords}

\section{Introduction}
Language is primarily spoken and only secondarily written. In fact, only about half of all living languages have
a formal writing system \cite{ethnologue}. 
Developments brought about by Large Language Models (LLMs), the language-specific performance of which is highly correlated with the amount of \textit{text} that exists for a language, further isolate the speakers of spoken-first languages with little-to-no text data.    
This points to the importance of Spoken Language Understanding (SLU), which focuses on making content-based predictions directly from speech utterances.  
Historically, however, SLU has been heavily influenced by task-oriented dialogue (ToD) \cite{budzianowski2018multiwoz,razumovskaia2022crossing}, resulting in datasets (\cite{speechcommands}, \cite{slurp}, \textit{inter alia}) built around ToD-specific tasks, intent classification and slot filling. These, arguably, require only superficial semantic capacities akin to detecting specific keywords. 
Other commonly used utterance-level SLU tasks like language identification \cite{speech-massive} or sentiment classification \cite{slue} are not content-based and probe largely phonetic or prosodic features of speech encoding. 
Unsurprisingly, the vast majority of SLU datasets are in English; moreover, the rare multilingual exceptions, 
such as the Minds14 dataset for intent classification \cite{minds14}, cover only high-resource languages. 

\begin{figure}[!t]
    \centering
    \includegraphics[width=1\columnwidth]{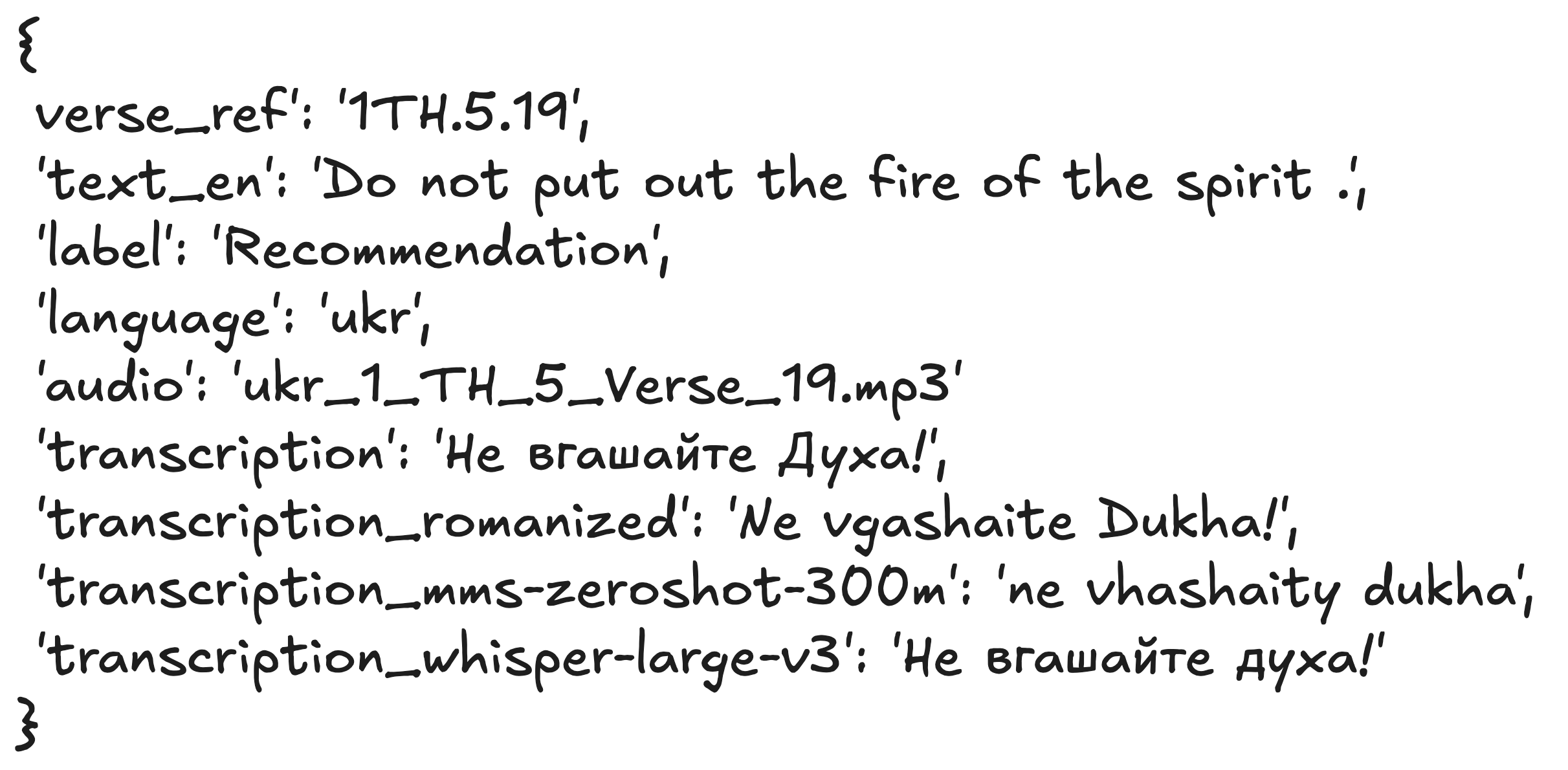}
    \caption{An instance of SpeechTaxi.}
    \label{fig:st-example}
\end{figure}

In this work, we introduce SpeechTaxi, an 80-hour multilingual SLU dataset for semantic speech classification that covers 28 languages, diverse both in terms of linguistic properties as well as ``resourceness''. We derive SpeechTaxi from Taxi1500 \cite{taxi1500}: it maps spoken bible verses into six classes (\textit{faith}, \textit{sin}, \textit{grace}, \textit{violence}, \textit{recommendation} and \textit{description}); Taxi1500 classes are abstract and semantically closely related (e.g., \textit{faith} vs. \textit{sin}), which makes the classification task challenging even for text-based models. An example instance from SpeechTaxi is shown in Figure \ref{fig:st-example}. 

With SpeechTaxi as a challenging multilingual SLU dataset in place, we leverage it to emprically compare the two most common SLU strategies in a multilingual setup, namely, (1) end-to-end classification (E2E) by means of fine-tuning a state-of-the-art massively multilingual speech encoder (MSE) and (2) a cascade approach (CA) in which we first transcribe speech with a state-of-the-art multilingual ASR model and then train a text-based classifier. On the one hand, we find that E2E is very effective in a monolingual scenario, when the MSE is fine-tuned on data from a single language: here E2E can significantly surpass CA. On the other hand, MSEs seem to poorly conduce cross-lingual transfer, with E2E lagging CA in multilingual training as well as in zero-shot cross-lingual transfer (ZS-XLT), i.e., when tested on languages unseen in training.               

Finally, we propose a viable cascading approach for low-resource spoken-only languages based on transcription to Romanized text. While recent efforts, e.g., the modular ASR models from \cite{scaling_1000}, significantly increased language coverage of ASR for low-resource languages, their reliance of transcribed speech for training makes it inapplicable to spoken-only languages.  
We simulate the scenario of semantic classification for spoken-only languages by conducting CA experiments for a set of 12 languages from our SpeechTaxi dataset that are not supported by Whisper \cite{whisper}, a state-of-the-art multilingual ASR model. We then compare a cascading approach based on original script, i.e., zero-shot transfer with a multilingual ASR model (ie., Whisper) against a language- and script-agnostic cascading approach with transcription to Romanized\footnote{Romanization generally maps non-Latin characters to their closest phonetic equivalents in the Latin alphabet \cite{uroman}. However, in contrast to proper phonetic representations (such as IPA), the phonological rules underlying those mappings and the resulting Romanized sequences are neither as detailed nor as expressive. Yet, they form a valuable approximation that sidesteps some of the practical implications that complicate working with Phonemes like (a) differing phoneme inventories or (b) mismatches between narrow and broad phonological transcriptions.} text. For the latter, we leverage MMS-Zeroshot (MMS-ZS) \cite{mms-uroman-asr}.

\section{SpeechTaxi}
\begin{figure}
    \centering
    \includegraphics[width=1\columnwidth]{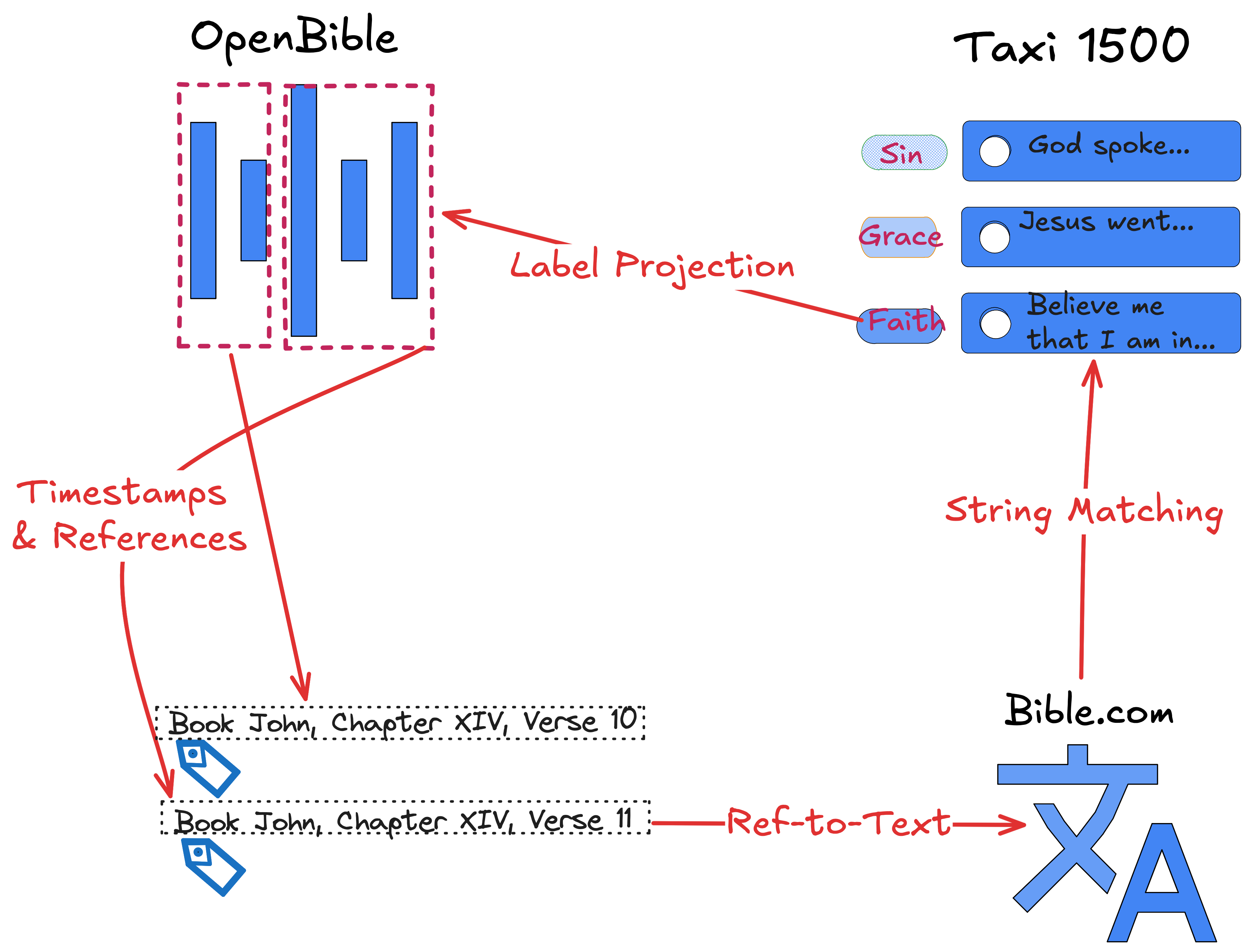}
    \caption{Illustration of SpeechTaxi creation from Taxi1500 and OpenBible data.}
    \label{fig:enter-label}
\end{figure}
\begin{table}[!t]
\tiny
\setlength{\tabcolsep}{10pt}
\centering
\label{table:dataset-summary}
\caption{SpeechTaxi Dataset Overview. \\ $|N|=$ Number of instances, $\Sigma=$ total audio length in hours, and  $\mu=$ average sample length in seconds.}
\resizebox{0.95\columnwidth}{!}{%
\begin{tabular}{|lllll|}
\hline
\textbf{Language} & \textbf{Script} & $\boldsymbol{|N|}$ & $\boldsymbol{\Sigma}$ & $\boldsymbol{\mu}$ \\ \hline
Vietnamese & Latin & 1076 & 2.6 & 8.7 \\
French & Latin & 1065 & 2.9 & 9.7 \\
Russian & Cyrillic & 1031 & 2.8 & 9.8 \\
Ukrainian & Cyrillic & 963 & 2.7 & 10.0 \\
North Ndebele & Latin & 949 & 2.9 & 11.0 \\
Yoruba & Latin & 932 & 2.8 & 10.6 \\
Kannada & Kannada & 930 & 3.1 & 12.2 \\
Akuapem Twi & Latin & 930 & 2.3 & 9.1 \\
Gujarati & Gujarati & 929 & 2.4 & 9.4 \\
Naga Pidgin & Latin & 929 & 3.2 & 12.4 \\
Urdu & Latin & 929 & 3.0 & 11.7 \\
Assamese & Bengali & 928 & 2.5 & 9.7 \\
Luo & Latin & 928 & 2.3 & 8.8 \\
Nahali & Devanagari & 928 & 2.3 & 9.1 \\
Panjabi & Gurmukhi & 928 & 2.6 & 10.2 \\
Tamil & Tamil & 927 & 2.8 & 10.9 \\
Asante Twi & Latin & 927 & 2.4 & 9.4 \\
Ewe & Latin & 924 & 3.4 & 13.4 \\
Marathi & Devanagari & 922 & 2.2 & 8.4 \\
Igbo & Latin & 921 & 3.0 & 11.5 \\
Haryanvi & Devanagari & 918 & 3.4 & 13.4 \\
Telugu & Telugu & 918 & 2.5 & 9.7 \\
Malayalam & Malayalam & 917 & 2.5 & 10.0 \\
Bhattiyali & Devanagari & 914 & 3.7 & 14.5 \\
Central Kurdish & Arabic & 907 & 2.8 & 11.0 \\
Pengo & Oriya & 892 & 3.8 & 15.5 \\
English & Latin & 724 & 2.0 & 9.7 \\
Lingala & Latin & 704 & 3.4 & 17.3 \\ \hline
\textbf{28} & \textbf{12} & \textbf{25890} & \textbf{78.3} & \textbf{11.0} \\ \hline
\end{tabular}%
}
\end{table}

We start from Taxi1500 \cite{taxi1500}, a dataset encompassing 1077 English Bible verses annotated with six (topical) labels; the multi-parallel nature of the Bible corpus \cite{mayer2014creating} is then leveraged to project the annotations onto the same Bible verses in other languages. To build SpeechTaxi, we too exploit the immense spread of the Bible: we gather the corresponding verses from Bible audiobooks in 28 languages and then project the Taxi1500 labels onto the audio. Figure \ref{fig:enter-label} visually summarizes the steps involved in creation of SpeechTaxi. 

\subsection{Alignment and label projection} 
The main challenge was to identify the exact locations of Taxi1500 verses (or variants thereof) within Bible audiobooks. 
To this end, we make use of two data sources that provide reliable verse alignments: OpenBible\footnote{\url{https://open.bible/about/}}, a platform that offers manually aligned Bible recordings for various languages (under permissive CC licenses) and the MASS dataset \cite{mass-dataset}, which provides verse alignments for a subset of the CMU-Wilderness dataset \cite{cmu-wilderness}. 

OpenBible provides chapter-level recordings with timestamps and bibliographical references for verses (e.g., \textit{Book of Matthew}, \textit{Chapter 1}, \textit{Verse 1}): we use these directly to split the chapter-level recordings into verse-level segments. 
Since OpenBible does not provide the verse text, we additionally leveraged \url{bible.com}, which couples bibliographical references for verses with verse text, to obtain transcriptions for our verse-level recordings. In cases where \url{bible.com} offered multiple translations for a verse, we additionally employed the alignment model from \cite{scaling_1000} to compute sequence-level alignment scores
between the verse-level recording and each of the texts to select the most likely transcription for the recording. This procedure yielded a corpus of verse-level audio-text pairs.\footnote{Of which we only use a small subset for SpeechTaxi and plan to release the remainder of the data as an ASR/TTS-dataset to facilitate research in multilingual speech technology beyond SLU.} Unlike the OpenBible, the MASS dataset already offers verse-level aligned audio-transcription pairs. 

Finally, we projected the Taxi1500 verse-level labels to our verse-level recordings by text-matching the transcriptions from our verse-level speech-text pairs against the text of the verse in Taxi1500.\footnote{We had to resort to text matching because the bibliographic references are not present in Taxi1500.}
Like \url{bible.com}, Taxi1500 also contains multiple translations for some languages, so we first collected all variants of each Taxi1500 verse; 
we then performed exhaustive all-to-all string comparisons between the Taxi1500 variants and our verse transcriptions and kept only the instances where we found exact string matches.\footnote{We also investigated fuzzy matching via token-level Jaccard distance with a threshold but found that, compared to exact matching, it did not significantly increase recall and led to reduced precision.}   

\subsection{Quality Assurance}
We carried out the following two-step procedure as the final quality assurance for SpeechTaxi instances. First, for the OpenBible data only, we used the CTC-alignment scores and manually set a threshold to filter out faulty alignments and only partially aligned audio-text pairs. Next, we randomly sampled 50 instances per language from the entire dataset and manually verified the correctness of the alignment between the audio, transcriptions (both from Taxi1500 and our data), the label, and the original English verse. For non-Latin scripts, we used Romanization to aid manual comparison and machine translation to validate consistency between the transcriptions and the original English verse. The second step revealed no errors among the instances sampled for manual inspection. 

\subsection{Dataset summary, splits, and availability}
Table \ref{table:dataset-summary} gives an overview of the final SpeechTaxi dataset.
As both data sources lack speaker-related metadata, we can't provide exact details, but since each language is represented by a single audiobook, we assume there is typically one speaker per language.
For compatibility and to prevent data leakage between training and test sets across languages, we adopted the same train, test, and validation splits as Taxi1500.

%
The CMU-Wilderness dataset is based on freely available but copyrighted audiobooks, which third parties may not redistribute: this also propagates to derivatves, i.e., the MASS dataset and the parts of SpeechTaxi built from it.\footnote{This portion of SpeechTaxi covers only French, English, and Russian.} We release the free portion of our dataset at \url{https://huggingface.co/datasets/LennartKeller/SpeechTaxi/}; we will also release code to replicate the copyrighted data.
\section{Experimental Setup}
\subsection{Models}

\sparagraph{Classification from speech encoding.} For the E2E approach, we employ two state-of-the-art massively multilingual MSEs:  MMS-1b \cite{scaling_1000} and XEUS \cite{xeus}. MMS-1b is a Wav2Vec2.0-style \cite{wav2vec20} model pretrained on a speech corpus encompassing 491K hours of speech in 1,406 languages from various sources.
XEUS was pretrained on an even larger speech corpus of roughly 1 million hours in 4,057 languages and uses (a) a different architecture than MMS (E-Branchformer \cite{e-branchformers} instead of Transformer \cite{att-is-all}) and (b) a more refined pretraining strategy that relies on (i) predicting phonetic pseudo-labels generated by a teacher MSE \cite{wavlabl-lm} for masked-out timesteps using cross-entropy loss and (ii) WavLM-style \cite{wavlm} addition of noise in conjunction with artificial reverberation of the model inputs (but not the label generator inputs) to enforce the separation of features related to recording setting from the actual speech content. 

\rparagraph{Cascade: transcription + classification from text encoding.} The cascaded setup (CA) couples a transcription model with a classifier based on a text encoder. For the former, we experiment with (1) transcription to the native script of respective languages, for which we leverage Whisper-large-v3 \cite{whisper} and (2) language-agnostic Romanized transcription, for which we resort to MMS-ZS \cite{mms-uroman-asr}. We use both models' default inference and decoding settings without any postprocessing of the output text.\footnote{In a truly language-agnostic setup, we rely on Whisper's built-in language detection component and do not provide the language token as a prefix.}

We then forward the transcribed text to one of the following two text encoders (TEs): Furina \cite{translico} and a Llama3 decoder \cite{llama3} converted into a bidirectional encoder model using the LLM2Vec procedure \cite{LLM2Vec}, hereafter referred to as LLM2Vec. Furina is a multilingual TE based on Glot500 \cite{glot500}, specifically adapted to process Romanized text by means of continued pretraining with an objective that aligns the representations of original and Romanized versions of text sequences. 
The LLM2Vec procedure converts autoregressive LLMs into powerful text encoders by (i) replacing the causal attention mask in favor of a bidirectional one and continued self-supervised pretraining with (ii) masked-next token prediction as well as (iii) sequence representation learning via SimCSE \cite{simcse}. 
We selected these TEs based on the following considerations. Furina is multilingual and adapted to Romanization, and thus can be seen as specialized for the output of a MMS-ZS which transcribes speech to Romanized text. In contrast, LLM2Vec is a derivative of one of the most potent general-purpose LLMs currently available, the performance of which is, however, heavily skewed towards English\footnote{For example, Llama3.1's pretraining corpus contains twice as much programming code as it does multilingual data (17\% vs 8\%).}; nonetheless, being derived from a potent LLM, LLM2Vec should be able to generalize well to new data distributions, largely unseen during pretraining (specifically, (Romanized) non-English transcriptions). In this sense, CA with LLM2Vec can indicate how effective CA may be for entirely unseen languages (e.g., as in the case of spoken-only languages).




\subsection{Training and Hyperparameters}
\sparagraph{Task-specific architecture}
To support a more direct comparison between the different models, in both E2E and CA, we refrain from using any specialized task-specific architectures and only add a simple softmax classifier on top of the encoder's output. 
To obtain sequence-level representations, we follow the model's standard recipes: we use the embedding of special sequence start/end token for TEs (\texttt{[BOS]} for Furina, \texttt{[EOS]} for LLM2Vec), and do the mean-pooling of ``audio-token representations'' for MSEs.

\rparagraph{Training details}
We train the models by minimizing the standard cross-entropy loss and, except for LLM2Vec, update all parameters during fine-tuning using mixed precision and either the Adam \cite{adam-optim} 
 or Adam(W) \cite{adamw-optim} optimizer (with weight decay $\lambda = 0.001$).
For LLM2Vec, full fine-tuning is prohibitively expensive due the model size, so we instead opt for a parameter-efficient fine-tuning approach and insert trainable LoRA parameters \cite{lora} (with rank $r=16$, alpha $\alpha=32$, and dropout of $p = 0.05$) at each feed-forward layer into the frozen base model 
To further alleviate the computational strain, we apply 4bit QLoRA-quantization\footnote{Only to encoder, the classifier remains in mixed 16-bit precision.} \cite{qlora} and use an 8bit-AdamW optimizer \cite{adamw-8bit}.
For both MSEs, we follow the standard practice and enable SpecAugment \cite{specaugment} during fine-tuning to improve convergence.
We train all models with an effective batch size of 16, a learning-rate schedule with a warmup for the first 10\% of the training, and linear decay for the remainder. With extensive hyperparameter search being prohibitively expensive, for each model we adopt the hyperparameter values reported in the original work. 
Table \ref{table:hparams} summarizes the values of key hyperparameters for all encoders.

\begin{table}[!t]
\caption{Hyperparameter Summary}
\tiny
\centering
\setlength\tabcolsep{2pt}
\begin{tabular}{|c|c|c|cc|}
\hline
\textbf{Model}                & \textbf{LR}      & \textbf{Optimizer}              & \multicolumn{2}{c|}{\textbf{No. Epochs}} \\ \cline{4-5} 
                              &                             &                                 & \textit{Single} & \textit{Joined}    \\ \hline
MMS-1b                        & $3 \times 10^{-5}$  & AdamW                          & 100            & 20                \\ \hline
XEUS                          & $2 \times 10^{-4}$  & AdamW                          & 100            & 20                \\ \hline
Furina                        & $1 \times 10^{-5}$  & Adam                           & 100            & 20                \\ \hline
LLM2Vec                       & $2 \times 10^{-4}$  & AdamW-8Bit                     & 40             & 15                \\ \hline
\end{tabular}
\label{table:hparams}
\end{table}



\subsection{Experiments}
We evaluate both E2E and CA for three different training configurations: 
\textbf{1)} \textit{In-Language}: we train (and evaluate) monolingual models, i.e., using training data of one language only;
\textbf{2)} \textit{All-Languages}: joint multilingual training on all 28 languages and evaluate in all languages; and 
\textbf{3)} \textit{Zero-shot cross-lingual transfer (ZS-XLT)}: we evaluate a monolingual model trained on data from a high-resource language, English or French, on all other SpeechTaxi languages.
\section{Results and Discussion}
\begin{table}[!t]
\label{table:all-results-averaged}
\centering  
\caption{Results of semantic speech classification on SpeechTaxi.}
\setlength\tabcolsep{3pt}
\resizebox{\columnwidth}{!}{%
\begin{tabular}{|c|cc|cccc|}
\hline
\multirow{3}{*}{} & \multicolumn{2}{c|}{\textbf{E2E}} & \multicolumn{4}{c|}{\textbf{CA}} \\ \cline{2-7} 
 & \multicolumn{1}{c|}{\multirow{2}{*}{\textbf{MMS-1b}}} & \multirow{2}{*}{\textbf{XEUS}} & \multicolumn{2}{c|}{\textbf{Furina}} & \multicolumn{2}{c|}{\textbf{LLM2Vec}} \\ \cline{4-7} 
 & \multicolumn{1}{c|}{} &  & \multicolumn{1}{c|}{\textbf{Orig. Script}} & \multicolumn{1}{c|}{\textbf{Romanized}} & \multicolumn{1}{c|}{\textbf{Orig. Script}} & \textbf{Romanized} \\ \hline
\textit{In-Language} & \multicolumn{1}{c|}{49.4} & 55.7 & \multicolumn{1}{c|}{51.5} & \multicolumn{1}{c|}{47.2} & \multicolumn{1}{c|}{50.9} & 50.3 \\
\textit{All-Languages} & \multicolumn{1}{c|}{48.8} & 43.7 & \multicolumn{1}{c|}{54.2} & \multicolumn{1}{c|}{54.6} & \multicolumn{1}{c|}{51.3} & 52.7 \\
\textit{ZS-XLT: Src=EN} & \multicolumn{1}{c|}{12.7} & 7.9 & \multicolumn{1}{c|}{34.6} & \multicolumn{1}{c|}{26.4} & \multicolumn{1}{c|}{18.6} & 10.5 \\
\textit{ZS-XLT: Src=FR} & \multicolumn{1}{c|}{14.2} & 11.0 & \multicolumn{1}{c|}{39.7} & \multicolumn{1}{c|}{28.5} & \multicolumn{1}{c|}{27.5} & 21.8 \\ \hline
\end{tabular}%
}
\vspace{-0.5em}
\end{table}

\begin{table}[!t]
\small
\setlength{\tabcolsep}{1pt}
\centering
\caption{Aggregate performance on two groups of languages: supported and unsupported by ASR (Whisper).}
\label{table:groupby-whisper}
\resizebox{\columnwidth}{!}{%
\begin{tabular}{|cc|cccc|cc|}
\hline
 & \multirow{2}{*}{} & \multicolumn{4}{c|}{\textbf{CA}} & \multicolumn{2}{c|}{\textbf{E2E}} \\ \cline{3-8} 
\textbf{} &  & \multicolumn{2}{c|}{\textbf{Furina}} & \multicolumn{2}{c|}{\textbf{LLM2Vec}} & \multicolumn{1}{c|}{\multirow{2}{*}{\textbf{MMS-1b}}} & \multirow{2}{*}{\textbf{XEUS}} \\ \cline{2-6}
\multicolumn{1}{|c|}{} & \textbf{Whisper ?} & \multicolumn{1}{c|}{\textbf{Orig. Script}} & \multicolumn{1}{c|}{\textbf{Romanized}} & \multicolumn{1}{c|}{\textbf{Orig. Script}} & \textbf{Romanized} & \multicolumn{1}{c|}{} &  \\ \hline
\multicolumn{1}{|c|}{\multirow{2}{*}{\textit{In-Language}}} & \textit{Yes} & \multicolumn{1}{c|}{59.3} & \multicolumn{1}{c|}{49.0} & \multicolumn{1}{c|}{57.4} & 53.8 & \multicolumn{1}{c|}{49.9} & 57.0 \\
\multicolumn{1}{|c|}{} & \textit{No} & \multicolumn{1}{c|}{41.2} & \multicolumn{1}{c|}{44.8} & \multicolumn{1}{c|}{42.2} & 45.6 & \multicolumn{1}{c|}{48.7} & 54.0 \\ \hline
\multicolumn{1}{|c|}{\multirow{2}{*}{\textit{All-Langauges}}} & \textit{Yes} & \multicolumn{1}{c|}{62.3} & \multicolumn{1}{c|}{57.2} & \multicolumn{1}{c|}{59.0} & 56.1 & \multicolumn{1}{c|}{49.2} & 44.0 \\
\multicolumn{1}{|c|}{} & \textit{No} & \multicolumn{1}{c|}{43.4} & \multicolumn{1}{c|}{51.0} & \multicolumn{1}{c|}{41.0} & 48.2 & \multicolumn{1}{c|}{48.3} & 43.2 \\ \hline
\multicolumn{1}{|c|}{\multirow{2}{*}{\textit{ZS-XLT SRC=Fr}}} & \textit{Yes} & \multicolumn{1}{c|}{50.7} & \multicolumn{1}{c|}{33.2} & \multicolumn{1}{c|}{35.6} & 25.2 & \multicolumn{1}{c|}{15.3} & 11.0 \\
\multicolumn{1}{|c|}{} & \textit{No} & \multicolumn{1}{c|}{25.9} & \multicolumn{1}{c|}{22.8} & \multicolumn{1}{c|}{17.3} & 17.5 & \multicolumn{1}{c|}{12.8} & 11.0 \\ \hline
\end{tabular}%
}
\end{table}

\begin{table}[!t]
\label{table:all-langauges-results-detailed}
\centering  
\caption{Detailed results for the All-Languages experiments. \textit{Italic}: languages not supported by Whisper. \textbf{Bold}: the best performance for a language.}
\setlength\tabcolsep{1pt}
\resizebox{\columnwidth}{!}{%
\begin{tabular}{|c|cc|cccc|}
\hline
                                   & \multicolumn{2}{c|}{\textbf{E2E}}                                                      & \multicolumn{4}{c|}{\textbf{CA}}                                                                                                                       \\ \hline
\multirow{2}{*}{\textbf{Language}} & \multicolumn{1}{c|}{\multirow{2}{*}{\textbf{MMS-1b}}} & \multirow{2}{*}{\textbf{XEUS}} & \multicolumn{2}{c|}{\textbf{Furina}}                                                 & \multicolumn{2}{c|}{\textbf{LLM2Vec}}                           \\ \cline{4-7} 
                                   & \multicolumn{1}{c|}{}                                 &                                & \multicolumn{1}{c|}{\textbf{Orig. Script}} & \multicolumn{1}{c|}{\textbf{Romanized}} & \multicolumn{1}{c|}{\textbf{Orig. Script}} & \textbf{Romanized} \\ \hline
\textit{Akuapem Twi}               & \multicolumn{1}{c|}{\bfseries 46.7}                   & 40.5                           & \multicolumn{1}{c|}{36.6}                  & \multicolumn{1}{c|}{42.6}               & \multicolumn{1}{c|}{33.4}                  & 38.1               \\
\textit{Asante Twi}                & \multicolumn{1}{c|}{\bfseries 47.4}                   & 42.2                           & \multicolumn{1}{c|}{28.5}                  & \multicolumn{1}{c|}{36.5}               & \multicolumn{1}{c|}{33.3}                  & 41.5               \\
Assamese                           & \multicolumn{1}{c|}{44.5}                             & 39.1                           & \multicolumn{1}{c|}{49.7}                  & \multicolumn{1}{c|}{50.5}               & \multicolumn{1}{c|}{52.3}                  & \bfseries 53.8     \\
\textit{Bhattiyali}                & \multicolumn{1}{c|}{40.7}                             & 40.2                           & \multicolumn{1}{c|}{55.8}                  & \multicolumn{1}{c|}{57.9}               & \multicolumn{1}{c|}{50.9}                  & \bfseries 61.5     \\
\textit{Central Kurdish}           & \multicolumn{1}{c|}{44.2}                             & 42.5                           & \multicolumn{1}{c|}{\bfseries 49.3}        & \multicolumn{1}{c|}{48.2}               & \multicolumn{1}{c|}{41.4}                  & 37.1               \\
English                            & \multicolumn{1}{c|}{48.5}                             & 52.3                           & \multicolumn{1}{c|}{79.3}                  & \multicolumn{1}{c|}{73.3}               & \multicolumn{1}{c|}{75.3}                  & \bfseries 80.1     \\
\textit{Ewe}                       & \multicolumn{1}{c|}{\bfseries 51.5}                   & 43.3                           & \multicolumn{1}{c|}{33.9}                  & \multicolumn{1}{c|}{47.2}               & \multicolumn{1}{c|}{40.9}                  & 51.2               \\
French                             & \multicolumn{1}{c|}{49.1}                             & 43.6                           & \multicolumn{1}{c|}{72.6}                  & \multicolumn{1}{c|}{67.6}               & \multicolumn{1}{c|}{\bfseries 74.1}        & 67.7               \\
Gujarati                           & \multicolumn{1}{c|}{50.9}                             & 49.9                           & \multicolumn{1}{c|}{\bfseries 67.6}        & \multicolumn{1}{c|}{55.3}               & \multicolumn{1}{c|}{60.8}                  & 59.9               \\
\textit{Haryanvi}                  & \multicolumn{1}{c|}{52.9}                             & 48.8                           & \multicolumn{1}{c|}{\bfseries 66.6}        & \multicolumn{1}{c|}{58.1}               & \multicolumn{1}{c|}{61.0}                  & 64.9               \\
\textit{Igbo}                      & \multicolumn{1}{c|}{43.5}                             & 38.4                           & \multicolumn{1}{c|}{38.4}                  & \multicolumn{1}{c|}{\bfseries 45.8}     & \multicolumn{1}{c|}{30.0}                  & 38.7               \\
Kannada                            & \multicolumn{1}{c|}{57.8}                             & 38.5                           & \multicolumn{1}{c|}{\bfseries 58.8}        & \multicolumn{1}{c|}{55.4}               & \multicolumn{1}{c|}{53.7}                  & 53.0               \\
Lingala                            & \multicolumn{1}{c|}{46.3}                             & 41.2                           & \multicolumn{1}{c|}{51.5}                  & \multicolumn{1}{c|}{\bfseries 65.7}     & \multicolumn{1}{c|}{39.0}                  & 47.6               \\
\textit{Luo}                       & \multicolumn{1}{c|}{\bfseries 50.3}                   & 42.6                           & \multicolumn{1}{c|}{43.6}                  & \multicolumn{1}{c|}{49.8}               & \multicolumn{1}{c|}{31.1}                  & 31.4               \\
Malayalam                          & \multicolumn{1}{c|}{44.8}                             & 36.4                           & \multicolumn{1}{c|}{\bfseries 46.7}        & \multicolumn{1}{c|}{42.5}               & \multicolumn{1}{c|}{45.2}                  & 40.5               \\
Marathi                            & \multicolumn{1}{c|}{55.7}                             & 51.9                           & \multicolumn{1}{c|}{66.0}                  & \multicolumn{1}{c|}{62.6}               & \multicolumn{1}{c|}{\bfseries 66.3}        & 55.4               \\
\textit{Naga Pidgin}               & \multicolumn{1}{c|}{55.6}                             & 50.4                           & \multicolumn{1}{c|}{43.7}                  & \multicolumn{1}{c|}{\bfseries 60.7}     & \multicolumn{1}{c|}{48.5}                  & 60.0               \\
\textit{Nahali}                    & \multicolumn{1}{c|}{53.7}                             & 48.2                           & \multicolumn{1}{c|}{51.5}                  & \multicolumn{1}{c|}{\bfseries 61.8}     & \multicolumn{1}{c|}{46.5}                  & 58.2               \\
\textit{North Ndebele}             & \multicolumn{1}{c|}{36.5}                             & 35.1                           & \multicolumn{1}{c|}{37.7}                  & \multicolumn{1}{c|}{\bfseries 47.5}     & \multicolumn{1}{c|}{33.8}                  & 43.3               \\
Panjabi                            & \multicolumn{1}{c|}{58.5}                             & 53.2                           & \multicolumn{1}{c|}{\bfseries 68.7}        & \multicolumn{1}{c|}{59.4}               & \multicolumn{1}{c|}{64.1}                  & 64.5               \\
\textit{Pengo}                     & \multicolumn{1}{c|}{\bfseries 56.6}                   & 46.6                           & \multicolumn{1}{c|}{35.7}                  & \multicolumn{1}{c|}{56.0}               & \multicolumn{1}{c|}{40.6}                  & 53.0               \\
Russian                            & \multicolumn{1}{c|}{45.7}                             & 44.2                           & \multicolumn{1}{c|}{\bfseries 73.2}        & \multicolumn{1}{c|}{56.5}               & \multicolumn{1}{c|}{72.7}                  & 61.2               \\
Tamil                              & \multicolumn{1}{c|}{46.4}                             & 39.8                           & \multicolumn{1}{c|}{\bfseries 70.5}        & \multicolumn{1}{c|}{51.5}               & \multicolumn{1}{c|}{58.7}                  & 48.3               \\
Telugu                             & \multicolumn{1}{c|}{45.7}                             & 43.5                           & \multicolumn{1}{c|}{47.1}                  & \multicolumn{1}{c|}{\bfseries 50.3}     & \multicolumn{1}{c|}{44.9}                  & 41.0               \\
Ukrainian                          & \multicolumn{1}{c|}{57.3}                             & 44.5                           & \multicolumn{1}{c|}{72.1}                  & \multicolumn{1}{c|}{60.9}               & \multicolumn{1}{c|}{\bfseries 73.4}        & 67.6               \\
Urdu                               & \multicolumn{1}{c|}{50.0}                             & 43.1                           & \multicolumn{1}{c|}{\bfseries 69.0}        & \multicolumn{1}{c|}{62.5}               & \multicolumn{1}{c|}{63.1}                  & 63.1               \\
Vietnamese                         & \multicolumn{1}{c|}{44.7}                             & 40.6                           & \multicolumn{1}{c|}{\bfseries 71.8}        & \multicolumn{1}{c|}{46.2}               & \multicolumn{1}{c|}{64.9}                  & 45.8               \\
Yoruba                             & \multicolumn{1}{c|}{40.8}                             & 41.4                           & \multicolumn{1}{c|}{32.5}                  & \multicolumn{1}{c|}{\bfseries 55.7}     & \multicolumn{1}{c|}{35.2}                  & 47.4               \\ \hline
Average                            & \multicolumn{1}{c|}{48.8}                             & 43.7                           & \multicolumn{1}{c|}{54.2}                  & \multicolumn{1}{c|}{\bfseries 54.6}     & \multicolumn{1}{c|}{51.3}                  & 52.7               \\ \hline
\end{tabular}%
}
\vspace{-1em}
\end{table}

Table \ref{table:all-results-averaged} reports the results in all three configurations, averaged across all 28 SpeechTaxi languages. For each experiment, the reported result is the average Macro-F1 performance over three different runs (i.e., with different random seeds). 

\rparagraph{In-Language}
In a monolingual setting, the E2E approach is surprisingly competitive, and XEUS even yielding the best overall performance. However, the picture becomes more nuanced when we take the ASR support into account, as shown in Table \ref{table:groupby-whisper}. Here we can observe two things: 1) On languages with ASR  (i.e., supported by Whisper), the Whisper+Furina cascade slightly outperforms XEUS. However, for low-resource languages without ASR support 
both MSEs substantially outperform all CA variants and, importantly (and unlike CA), perform only slightly worse than for high-resource languages (i.e., those supported by Whisper).  
Regarding the choice of a TE for CA, Furina slightly outperforms LLM2Vec when operating on original script produced by Whisper. On Romanized transcriptions provided by MMS-ZS, however, LLM2Vec outperforms Furina: 
looking at Table \ref{table:groupby-whisper}, we notice that the difference originates from languages 
 actually supported by Whisper; for these, Furina exhibits a 10-point performance drop when switching from script to Romanized transcription, whereas the drop is much smaller for LLM2Vec (-3.4\% ).  
This finding is surprising, considering Furina's explicit specialization for Romanized text. 

\rparagraph{All-Languages}
While E2E is the most robust approach for obtaining good monolingual models, the picture drastically changes in the \textit{All-Languages} setup: while MSEs (and in particular XEUS) suffer from joint multilingual training, all four CA variants benefit from it. 
Here the language-agnostic Romanization-based CA  with Furina  not only obtains the best overall performance 
but also benefits the most from the additional data (improvement of +7.4 points over its respective In-Language performance): this suggests that Romanization facilitates cross-lingual generalization.



Table \ref{table:all-langauges-results-detailed} details the \textit{All-Languages} performance for all 28 SpeechTaxi languages. 
Interestingly, We see that the E2E approach (with MMS1-1b) outperforms CA on four languages (\textit{Akuapem Twi}, \textit{Asante Twi}, \textit{Luo}, and \textit{Pengo}), all of which are very low-resource. Even more interesting is the comparison of performance variance across languages between E2E and CA: While MSEs exhibit a more uniform performance, CA (especially based on original script) shows much better performance for high-resource than for low-resource languages.     
This might result from the knowledge mSEs acquire during pretraining, which is predominantly phonetic \cite{se-features-phonetic}. Thus, even ``prolonged exposure'' to a high-resource language in pretraining does not result in (much) better semantic capabilities for that language. In contrast, the symbolic nature of text makes it easier for multilingual TEs to ``overfit'' to languages that are much more represented in pretraining.

\rparagraph{ZS-XLT}
Cross-lingual transfer results compromise the promising results of E2E from the monolingual track. Compared to all CA variants, MSEs' ability to generalize from a single (high-resource) source language to many other is severely limited, irrespective of the source language (English or French). Taking a language's compatibility with Whisper as a proxy for its ``resourcesness'' (see Table \ref{table:groupby-whisper}), we see that ZS-XLT with E2E suffers for both high- and low-resource languages alike, with MMS-1b being only marginally better on languages that Whisper supports. 
Whisper's script-based transcription yields better CA performance than the Romanized transcription of MMS-ZS: we believe that this is merely because the TEs are better at processing script, since they have been pretrained on large script (and not Romanized) corpora; this, combined with results from All-Languages, suggests that at least some In-Language Romanized training data is required for an effective cross-lingual transfer.   
Still, even the best ZS-XLT performance falls dramatically behind the respective In-Language performance: this is to be primarily attributed to low-resource languages unsupported by Whisper (see again \ref{table:groupby-whisper}). 

\rparagraph{Language-agnostic CA}
Combining the results from all three experimental tracks, the Romanization-based CA emerges as an effective way to enable SLU for languages without ASR support, especially if task-specific training data is available in multiple languages. However, it is not a one-size-fits-all solution since, in monolingual settings, XEUS is the better option for low-resource languages (without ASR support, Table \ref{table:groupby-whisper}), and CA with proper ASR remains the best choice for ASR-supported languages. 
\section{Conclusion}

We introduced SpeechTaxi, an 80-hour multilingual SLU dataset for semantic speech classification across 28 diverse languages.
Using SpeechTaxi, we compared the two dominant SLU paradigms, End-to-End speech-based classification (E2E) and Cascade of ASR and text-based classification (CA), across various settings, leveraging state-of-the-art text and speech models.
Through extensive experiments in three different training scenarios, we found that while CA remains the most versatile option for SLU, modern massively multilingual speech encoders can be competitive in monolingual settings, particularly for low-resource languages with limited ASR support. However when training data is available in multiple languages, CA---based either on native script transcription (via Whisper) for high-resource languages or Romanized text transcription for low-resource languages without ASR support---remains the best performing approach for semantic speech classification.



\clearpage
\bibliographystyle{IEEEtran}
\bibliography{main}

\begin{thebibliography}{10}
\providecommand{\url}[1]{#1}
\csname url@samestyle\endcsname
\providecommand{\newblock}{\relax}
\providecommand{\bibinfo}[2]{#2}
\providecommand{\BIBentrySTDinterwordspacing}{\spaceskip=0pt\relax}
\providecommand{\BIBentryALTinterwordstretchfactor}{4}
\providecommand{\BIBentryALTinterwordspacing}{\spaceskip=\fontdimen2\font plus
\BIBentryALTinterwordstretchfactor\fontdimen3\font minus \fontdimen4\font\relax}
\providecommand{\BIBforeignlanguage}[2]{{%
\expandafter\ifx\csname l@#1\endcsname\relax
\typeout{** WARNING: IEEEtran.bst: No hyphenation pattern has been}%
\typeout{** loaded for the language `#1'. Using the pattern for}%
\typeout{** the default language instead.}%
\else
\language=\csname l@#1\endcsname
\fi
#2}}
\providecommand{\BIBdecl}{\relax}
\BIBdecl

\bibitem{ethnologue}
Ethnologue, ``{H}ow many languages in the world are unwritten? --- web.archive.org,'' \url{https://web.archive.org/web/20170227130129/https://www.ethnologue.com/enterprise-faq/how-many-languages-world-are-unwritten-0}, 2017.

\bibitem{budzianowski2018multiwoz}
P.~Budzianowski, T.-H. Wen, B.-H. Tseng, I.~Casanueva, S.~Ultes, O.~Ramadan, and M.~Gasic, ``Multiwoz-a large-scale multi-domain wizard-of-oz dataset for task-oriented dialogue modelling,'' in \emph{Proceedings of the 2018 Conference on Empirical Methods in Natural Language Processing}, 2018, pp. 5016--5026.

\bibitem{razumovskaia2022crossing}
E.~Razumovskaia, G.~Glavas, O.~Majewska, E.~M. Ponti, A.~Korhonen, and I.~Vulic, ``Crossing the conversational chasm: A primer on natural language processing for multilingual task-oriented dialogue systems,'' \emph{Journal of Artificial Intelligence Research}, vol.~74, pp. 1351--1402, 2022.

\bibitem{speechcommands}
P.~{Warden}, ``{Speech Commands: A Dataset for Limited-Vocabulary Speech Recognition},'' \emph{ArXiv e-prints}, Apr. 2018.

\bibitem{slurp}
E.~Bastianelli, A.~Vanzo, P.~Swietojanski, and V.~Rieser, ``{SLURP: A Spoken Language Understanding Resource Package},'' in \emph{{Proceedings of the 2020 Conference on Empirical Methods in Natural Language Processing (EMNLP)}}, 2020.

\bibitem{speech-massive}
B.~Lee, I.~Calapodescu, M.~Gaido, M.~Negri, and L.~Besacier, ``Speech-massive: A multilingual speech dataset for slu and beyond,'' in \emph{Interspeech 2024}, 2024, pp. 817--821.

\bibitem{slue}
S.~Shon, A.~Pasad, F.~Wu, P.~Brusco, Y.~Artzi, K.~Livescu, and K.~J. Han, ``Slue: New benchmark tasks for spoken language understanding evaluation on natural speech,'' in \emph{ICASSP 2022-2022 IEEE International Conference on Acoustics, Speech and Signal Processing (ICASSP)}.\hskip 1em plus 0.5em minus 0.4em\relax IEEE, 2022, pp. 7927--7931.

\bibitem{minds14}
D.~Gerz, P.~Su, R.~Kusztos, A.~Mondal, M.~Lis, E.~Singhal, N.~Mrksic, T.~Wen, and I.~Vulic, ``Multilingual and cross-lingual intent detection from spoken data,'' \emph{CoRR}, vol. abs/2104.08524, 2021.

\bibitem{taxi1500}
C.~Ma, A.~Googhari, H.~Ye, E.~Asgari, and H.~Schütze, ``Taxi1500: A multilingual dataset for text classification in 1500 languages,'' 2023.

\bibitem{scaling_1000}
V.~Pratap, A.~Tjandra, B.~Shi, P.~Tomasello, A.~Babu, S.~Kundu, A.~Elkahky, Z.~Ni, A.~Vyas, M.~Fazel-Zarandi, A.~Baevski, Y.~Adi, X.~Zhang, W.-N. Hsu, A.~Conneau, and M.~Auli, ``Scaling speech technology to 1,000+ languages,'' \emph{Journal of Machine Learning Research}, vol.~25, no.~97, pp. 1--52, 2024.

\bibitem{whisper}
A.~Radford, J.~W. Kim, T.~Xu, G.~Brockman, C.~Mcleavey, and I.~Sutskever, ``Robust speech recognition via large-scale weak supervision,'' in \emph{Proceedings of the 40th International Conference on Machine Learning}, ser. Proceedings of Machine Learning Research, A.~Krause, E.~Brunskill, K.~Cho, B.~Engelhardt, S.~Sabato, and J.~Scarlett, Eds., vol. 202.\hskip 1em plus 0.5em minus 0.4em\relax PMLR, 23--29 Jul 2023, pp. 28\,492--28\,518.

\bibitem{uroman}
U.~Hermjakob, J.~May, and K.~Knight, ``Out-of-the-box universal {R}omanization tool uroman,'' in \emph{Proceedings of {ACL} 2018, System Demonstrations}, F.~Liu and T.~Solorio, Eds.\hskip 1em plus 0.5em minus 0.4em\relax Melbourne, Australia: Association for Computational Linguistics, Jul. 2018, pp. 13--18.

\bibitem{mms-uroman-asr}
J.~Zhao, V.~Pratap, and M.~Auli, ``Scaling a simple approach to zero-shot speech recognition,'' 2024.

\bibitem{mayer2014creating}
T.~Mayer and M.~Cysouw, ``Creating a massively parallel bible corpus,'' \emph{Oceania}, vol. 135, no. 273, p.~40, 2014.

\bibitem{mass-dataset}
M.~Zanon~Boito, W.~Havard, M.~Garnerin, {\'E}.~Le~Ferrand, and L.~Besacier, ``\BIBforeignlanguage{English}{{M}a{SS}: A large and clean multilingual corpus of sentence-aligned spoken utterances extracted from the {B}ible},'' in \emph{\BIBforeignlanguage{English}{Proceedings of the Twelfth Language Resources and Evaluation Conference}}, N.~Calzolari, F.~B{\'e}chet, P.~Blache, K.~Choukri, C.~Cieri, T.~Declerck, S.~Goggi, H.~Isahara, B.~Maegaard, J.~Mariani, H.~Mazo, A.~Moreno, J.~Odijk, and S.~Piperidis, Eds.\hskip 1em plus 0.5em minus 0.4em\relax Marseille, France: European Language Resources Association, May 2020, pp. 6486--6493.

\bibitem{cmu-wilderness}
A.~W. Black, ``Cmu wilderness multilingual speech dataset,'' in \emph{ICASSP 2019 - 2019 IEEE International Conference on Acoustics, Speech and Signal Processing (ICASSP)}, 2019, pp. 5971--5975.

\bibitem{xeus}
W.~Chen, W.~Zhang, Y.~Peng, X.~Li, J.~Tian, J.~Shi, X.~Chang, S.~Maiti, K.~Livescu, and S.~Watanabe, ``Towards robust speech representation learning for thousands of languages,'' 2024.

\bibitem{wav2vec20}
A.~Baevski, Y.~Zhou, A.~Mohamed, and M.~Auli, ``wav2vec 2.0: A framework for self-supervised learning of speech representations,'' in \emph{Advances in Neural Information Processing Systems}, H.~Larochelle, M.~Ranzato, R.~Hadsell, M.~Balcan, and H.~Lin, Eds., vol.~33.\hskip 1em plus 0.5em minus 0.4em\relax Curran Associates, Inc., 2020, pp. 12\,449--12\,460.

\bibitem{e-branchformers}
A.~Gulati, J.~Qin, C.-C. Chiu, N.~Parmar, Y.~Zhang, J.~Yu, W.~Han, S.~Wang, Z.~Zhang, Y.~Wu, and R.~Pang, ``Conformer: Convolution-augmented transformer for speech recognition.'' in \emph{INTERSPEECH}, H.~Meng, B.~Xu, and T.~F. Zheng, Eds.\hskip 1em plus 0.5em minus 0.4em\relax ISCA, 2020, pp. 5036--5040.

\bibitem{att-is-all}
A.~Vaswani, N.~Shazeer, N.~Parmar, J.~Uszkoreit, L.~Jones, A.~N. Gomez, L.~u. Kaiser, and I.~Polosukhin, ``Attention is all you need,'' in \emph{Advances in Neural Information Processing Systems}, I.~Guyon, U.~V. Luxburg, S.~Bengio, H.~Wallach, R.~Fergus, S.~Vishwanathan, and R.~Garnett, Eds., vol.~30.\hskip 1em plus 0.5em minus 0.4em\relax Curran Associates, Inc., 2017.

\bibitem{wavlabl-lm}
W.~Chen, J.~Shi, B.~Yan, D.~Berrebbi, W.~Zhang, Y.~Peng, X.~Chang, S.~Maiti, and S.~Watanabe, ``Joint prediction and denoising for large-scale multilingual self-supervised learning,'' in \emph{2023 IEEE Automatic Speech Recognition and Understanding Workshop (ASRU)}, 2023, pp. 1--8.

\bibitem{wavlm}
S.~Chen, C.~Wang, Z.~Chen, Y.~Wu, S.~Liu, Z.~Chen, J.~Li, N.~Kanda, T.~Yoshioka, X.~Xiao, J.~Wu, L.~Zhou, S.~Ren, Y.~Qian, Y.~Qian, J.~Wu, M.~Zeng, and F.~Wei, ``Wavlm: Large-scale self-supervised pre-training for full stack speech processing,'' \emph{CoRR}, vol. abs/2110.13900, 2021.

\bibitem{translico}
Y.~Liu, C.~Ma, H.~Ye, and H.~Schuetze, ``{T}ransli{C}o: A contrastive learning framework to address the script barrier in multilingual pretrained language models,'' in \emph{Proceedings of the 62nd Annual Meeting of the Association for Computational Linguistics (Volume 1: Long Papers)}, L.-W. Ku, A.~Martins, and V.~Srikumar, Eds.\hskip 1em plus 0.5em minus 0.4em\relax Bangkok, Thailand: Association for Computational Linguistics, Aug. 2024, pp. 2476--2499.

\bibitem{llama3}
Llama-Team, ``The llama 3 herd of models,'' 2024.

\bibitem{LLM2Vec}
P.~BehnamGhader, V.~Adlakha, M.~Mosbach, D.~Bahdanau, N.~Chapados, and S.~Reddy, ``{LLM}2vec: Large language models are secretly powerful text encoders,'' in \emph{First Conference on Language Modeling}, 2024.

\bibitem{glot500}
A.~ImaniGooghari, P.~Lin, A.~H. Kargaran, S.~Severini, M.~Jalili~Sabet, N.~Kassner, C.~Ma, H.~Schmid, A.~Martins, F.~Yvon, and H.~Sch{\"u}tze, ``Glot500: Scaling multilingual corpora and language models to 500 languages,'' in \emph{Proceedings of the 61st Annual Meeting of the Association for Computational Linguistics (Volume 1: Long Papers)}, A.~Rogers, J.~Boyd-Graber, and N.~Okazaki, Eds.\hskip 1em plus 0.5em minus 0.4em\relax Toronto, Canada: Association for Computational Linguistics, Jul. 2023, pp. 1082--1117.

\bibitem{simcse}
T.~Gao, X.~Yao, and D.~Chen, ``{S}im{CSE}: Simple contrastive learning of sentence embeddings,'' in \emph{Proceedings of the 2021 Conference on Empirical Methods in Natural Language Processing}, M.-F. Moens, X.~Huang, L.~Specia, and S.~W.-t. Yih, Eds.\hskip 1em plus 0.5em minus 0.4em\relax Online and Punta Cana, Dominican Republic: Association for Computational Linguistics, Nov. 2021, pp. 6894--6910.

\bibitem{adam-optim}
D.~Kingma and J.~Ba, ``Adam: A method for stochastic optimization,'' in \emph{International Conference on Learning Representations (ICLR)}, San Diega, CA, USA, 2015.

\bibitem{adamw-optim}
I.~Loshchilov and F.~Hutter, ``Decoupled weight decay regularization,'' in \emph{International Conference on Learning Representations}, 2019.

\bibitem{lora}
E.~J. Hu, yelong shen, P.~Wallis, Z.~Allen-Zhu, Y.~Li, S.~Wang, L.~Wang, and W.~Chen, ``Lo{RA}: Low-rank adaptation of large language models,'' in \emph{International Conference on Learning Representations}, 2022.

\bibitem{qlora}
T.~Dettmers, A.~Pagnoni, A.~Holtzman, and L.~Zettlemoyer, ``{QL}o{RA}: Efficient finetuning of quantized {LLM}s,'' in \emph{Thirty-seventh Conference on Neural Information Processing Systems}, 2023.

\bibitem{adamw-8bit}
T.~Dettmers, M.~Lewis, S.~Shleifer, and L.~Zettlemoyer, ``8-bit optimizers via block-wise quantization,'' \emph{CoRR}, vol. abs/2110.02861, 2021.

\bibitem{specaugment}
D.~S. Park, W.~Chan, Y.~Zhang, C.-C. Chiu, B.~Zoph, E.~D. Cubuk, and Q.~V. Le, ``Specaugment: A simple data augmentation method for automatic speech recognition,'' in \emph{Interspeech 2019}, 2019, pp. 2613--2617.

\bibitem{se-features-phonetic}
K.~Choi, A.~Pasad, T.~Nakamura, S.~Fukayama, K.~Livescu, and S.~Watanabe, ``Self-supervised speech representations are more phonetic than semantic,'' in \emph{Interspeech 2024}, 2024, pp. 4578--4582.

\end{thebibliography}
\vspace{12pt}

\end{document}